\documentclass[10pt,journal,compsoc]{IEEEtran}

\usepackage{algorithm}
\usepackage{algorithmic}
\usepackage{amsfonts}
\usepackage{amsmath}
\usepackage{graphicx}
\usepackage{epstopdf}
\usepackage{tabularx}
\usepackage{color}
\usepackage{caption}
\usepackage{subcaption}
\usepackage{adjustbox}
\graphicspath{ {./figure/} }

\ifCLASSOPTIONcompsoc
  \usepackage[nocompress]{cite}
\else
  \usepackage{cite}
\fi
\ifCLASSINFOpdf
\else
\fi
\begin{document}

\title{Diversified  Hidden Markov Models \\for Sequential Labeling}

\author{Maoying~Qiao,~\IEEEmembership{Student Member,~IEEE,}
        Wei~Bian,~\IEEEmembership{Member,~IEEE,}\\
	 Richard~Yi~Da~Xu,~
        and~Dacheng~Tao,~\IEEEmembership{Fellow,~IEEE}
\IEEEcompsocitemizethanks{\IEEEcompsocthanksitem M. Qiao, W. Bian and D. Tao are with
the Centre for Quantum Computation \& Intelligent Systems and the Faculty of Engineering and Information Technology,
University of Technology, Sydney,
81 Broadway Street, Ultimo,
NSW 2007, Australia. \protect\\
R. Y. D. Xu is with the Faculty of Engineering and Information Technology,
University of Technology, Sydney,
81 Broadway Street, Ultimo,
NSW 2007, Australia. \protect\\
E-mail: maoying.qiao@student.uts.edu.au,
\{wei.bian, yida.xu, dacheng.tao\}@uts.edu.au
}
\thanks{\textcopyright 2015 IEEE. Personal use of this material is permitted. Permission from IEEE must be obtained for all other users, in any current or future media, including reprinting/republishing this material for advertising or promotional purposes, creating new collective works, for resale or redistribution to servers or lists, or reuse of any copyrighted component of this work in other works.}
}

\markboth{IEEE Transactions on Knowledge and Data Engineering}%
{Shell \MakeLowercase{\textit{et al.}}: Bare Advanced Demo of IEEEtran.cls for Journals}
%

\IEEEtitleabstractindextext{%
\begin{abstract}
Labeling of sequential data is a prevalent meta-problem for a wide range of real world applications. While the first-order Hidden Markov Models (HMM) provides a fundamental approach for unsupervised sequential labeling, the basic model does not show satisfying performance when it is directly applied to real world problems, such as part-of-speech tagging (PoS tagging) and optical character recognition (OCR). Aiming at improving performance, important extensions of HMM have been proposed in the literatures. One of the common key features in these extensions is the incorporation of proper prior information. In this paper, we propose a new extension of HMM, termed diversified Hidden Markov Models (dHMM), which utilizes a diversity-encouraging prior over the state-transition probabilities and thus facilitates more dynamic sequential labellings. Specifically, the diversity is modeled by a continuous determinantal point process prior, which we apply to both unsupervised and supervised scenarios. Learning and inference algorithms for dHMM are derived. Empirical evaluations on benchmark datasets for unsupervised PoS tagging and supervised OCR confirmed the effectiveness of dHMM, with competitive performance to the state-of-the-art.
\end{abstract}

}

\maketitle

\IEEEdisplaynontitleabstractindextext

\IEEEpeerreviewmaketitle

\ifCLASSOPTIONcompsoc
\IEEEraisesectionheading{\section{Introduction}\label{sec:introduction}}
\else
\section{Introduction}
\label{sec:introduction}
\fi

\IEEEPARstart{S}{equential} labeling is an important meta-problem in many real world applications, including natural language processing (NLP) tasks \cite{HMMNLP1990} \cite{HMMNER2012}, video analysis \cite{HMMVideo2005}\cite{HMMVideoFace2003}, protein secondary structure\cite{HMMProtein2001} \cite{HMMProteinStructure1992}. It has received considerable attentions in the past years. One of the fundamental models for sequential labeling is HMM, which assumes a ‘chain’ of discrete-valued latent states and each of them depends only on the immediate neighboring states. Conditioning on this latent chain, the observations are probabilistically independent. Take PoS tagging task from NLP as an example, speech tags (NNS-Noun, plural; MD-Modal; etc) are discrete-valued latent state, while words (directors; are; etc) are observations.

However, the parameter-learning task in the classical HMM implemented with expectation-maximization (EM) algorithm performs unsatisfactorily in unsupervised setting for sequential labeling \cite{WhynotEM07}. A key reason for this drawback is the well-known fact that maximum likelihood estimation (MLE) with mixture parameters has the tendency to converge towards a singular estimate at the boundary of the parameter space \cite{unsupervisedlearningMM00}\cite{bishop2006pattern}, no matter how the observations are actually distributed (e.g. normally distributed or askew distributed). With improperly estimated parameters, the performance of inference of the latent states can be severely unsatisfied. Besides, the identifiability of parameters is another issue for HMM parameter learning with MLE implementation.

Therefore, a penalized MLE with properly chosen prior distribution over parameters is essential for the practical applications of HMM. For example, smoothing penalty \cite{QinWang2005} and sparse penalty  \cite{sparseHMM2012} \cite{Zechao2014} \cite{Shangming2014} \cite{Mingsheng2014} are two popular priors over either transition distribution parameters or emission distribution parameters for sequential labeling.

Different from the early works, we have explored the usage of diversity prior over a joint distribution of rows of HMM’s transition matrix, in order to make these transition distributions more distinct when decoding the sequential latent states.
In the cases, where the transition probabilities become similar, an HMM model approaches to a static mixture model.
We can We can understand this intuition by considering an extreme case where each rows of a transition probability matrix are identical. This leads to the same state transitions regardless of the state we are currently in. Suppose we have a  k-state HMM, parameterised by $(\pi,A,B)$, i.e., the initial probability $\pi$, transition probability matrix $A$ and emission probability $B$, if the rows of $A$ are identical and given by vector $a$, the joint probability of the hidden states and observations over a sequence of length T can be calculated as:
\begin{align}
P(X,Y|\pi,A,B) &= P(x_1|\pi)\prod_{t=2}^T P(x_t|A(x_{t-1},:)) P(y_t|x_t,B) \nonumber \\
&= P(x_1|\pi)\prod_{t=2}^T P(x_t|a) P(y_t|x_t,B) \nonumber
\end{align}
It can be seen that the joint probability becomes an independent product of variables at individual time point, and thus the HMM model becomes a static mixture model, i.e., the data become exchangeable. In contrast, a prior that encouraging diversity is able to reserve the dynamics property of HMMs.
To the best of our knowledge, this is the first paper to apply diversity prior over HMM parameters.
We do so by incorporating a recently introduced Determinantal Point Processes (DPP) \cite{DPP12} methodology, which essentially defines a Probability Mass Function (PMF) that assigns higher probability to a diverse subset of data. Inspired by the work of \cite{JamesZou2012}, we propose a diversified HMM (dHMM) by extending the basic HMM with determinantal-driven diversity.


Specifically, our contributes can be summarized in the following:

\begin{itemize}
\item
We extend the HMM to dHMM by incorporating a diversity-encouraging prior over transition distributions, with which we intend to mitigate the problem of singular estimate in HMM.

\item The use of a prior does not change the E-step of the EM algorithm in a fundamental way. However, for the M-step of the estimation of dHMM parameters, we derive a new formulation to incorporate the continuous DPP prior over the transition probabilities.

\item We demonstrate the effectiveness of dHMM under both the unsupervised and supervised settings by applying dHMM to benchmark sequential labeling problems, including: part-of-speech tagging (PoS tagging) and optical character recognition (OCR).
\end{itemize}

The rest of this paper is organized as follows.
Section \ref{sec:relatedwork} reviews related literatures on both progresses of statistical HMM model for sequential labeling and recent development on DPP.
Section \ref{sec:dHMM} introduces our proposed model in detail. We briefly review both continuous Determinantal Point Processes (DPP) based on probability kernel and basic Hidden Markov Models (HMM). How diversity-encouraging prior is encoded into HMM and how the induced Maximum A Posterior (MAP) objective problem is solved are also presented in details. Both simulated and real-world experiments are conducted in section \ref{sec:exps}. Finally, conclusion is discussed in Section  \ref{sec:conclusion}.

\section{Related Work}
\label{sec:relatedwork}
Our work utilized two fundamental machine learning frameworks: HMM and DPP.
On one hand, HMM is a fundamental statistic model for modeling sequential dataset and of significant importance in many other fields, from speech recognition\cite{ClassicalHMM1989}, handwriting recognition \cite{HMMHandwriting1996}, video analysis \cite{HMMVideoFace2003}, gesture recognition \cite{sparseHMM2012}, gene sequence prediction \cite{HMMProtein2001}, to optical character recognition (OCR) \cite{HMMOCR2006} and part-of-speech tagging (PoS tagging)) \cite{infiniteHHPOS09}. On the other hand,
DPP is a probabilistic model of repulsion in quantum physics area and recently introduced in machine learning area. Since it has closed-form solution for inference, recently, lots of its extensions are developed. This section presents a review of previous work in both DPP and HMM.

\subsection{HMM extensions with priors}
We briefly review the development of HMM for sequential labeling.
It is well known that the HMM parameters contain three parts: (1) initial state distribution, (2) one transition distributions and (3) parameters associated with emission distributions, discrete or continuous. Various extensions of HMM have been proposed by incorporating proper prior information onto either one part or all three parts of these parameters.

Supervised sparse HMM \cite{sparseHMM2012} was proposed to improve the expressive power for sequential surgical gesture classification and skill evaluation. It assumes that the emission distributions sparsely and linearly constitute elements from dictionary of basic surgical motions, no matter the observations are discrete, Gaussian or factor analyzed. Training dataset is needed for dictionary learning for each gesture together with an HMM grammar describing the transitions among different gestures. With learned dictionaries and grammar, the testing motion data is represented and classified.

Supervised large margin continuous density HMM (CD-HMM) for automatic speech recognition was proposed by Sha and Saul in \cite{LMHMM2006}.
The real-valued observations (such as acoustic feature vectors) are modeled through Gaussian mixture models. Inspired by support vector machines, margin maximization is applied as training objective function which is defined over a parameter space of positive semidefinite matrices. This optimization problem can be solved efficiently with simple gradient-based methods.

Unsupervised learning is a more difficult but important problem, as it eliminates the need for expensive manual annotation.
It was demonstrated from the work of  \cite{QinWang2005} that, smoothing HMM parameters can achieve significant improvements for PoS tagging. Two strategies have been applied: the first one is to smooth the emission distributions by computing observed word similarities. The second one is to specify a stationary distribution for hidden states to constrain the transition distributions.

Unsupervised sparse HMM based on Bayesian framework also has been explored in several literatures. Unlike supervised sparse HMM \cite{sparseHMM2012}, the emission distributions are learned from sparse representation, unsupervised sparse models add priors on transition distributions. \cite{sparsenessMHMM2007} introduces a negative Dirichlet prior on the transition distribution, which strongly encourages sparseness of the model. Then, maximum a posteriori (MAP) probability estimation of HMM parameters is devised under a modified Expectation Maximization algorithm. Manuele et al. evaluate the proposed technique on a 2D shape classification task.
In \cite{BayesianSharon05}, for PoS tagging, rather than performing MAP parameters estimation first then followed by inferring hidden states, Goldwater and Griffiths directly identify a distribution over latent variables, without ever fixing particular values for the model parameters. This is achieved by integrating over all possible values of parameters under a Bayesian approach. The integrating over parameters space permits the usage of linguistic appropriate priors. For example, the symmetric Dirichlet prior may prefer equally, uniform or sparse multinomial distributions according to different settings of its hyper-parameters.

There exist other important extensions of HMM for determining the number of hidden states 
which is a key parameter in all clustering tasks.
Non-parametric bayesian method is one popular solution for this problem. For instance, Guojun Qi et al. \cite{GuojunWSDM13} applied hierarchical dirichlet processes prior over transition matrix to model the number evolution of the dynamic community structures.
Here we fix this parameter from experience and we refer to 
\cite{infiniteHHPOS09} \cite{teh2006hierarchical} for interested readers.


From these previous works, proper prior information \cite{Fang2015}
encoded into HMM leads to visible performance increment.
Either smooth prior or sparse prior is somewhat reasonable from technical view.
From intuitive view, different states would have different transition distributions, otherwise, HMM will finally fall to a 'static' mixture model. Diversity-encouraging prior is reasonable and also natural to many real-world sequential applications. In next subsection, we review DPP, which is an elegant statistical models of diversity, and its recent developments.

\subsection{DPP models of diversity}

Determinantal Point Processes (DPP) provide a probability measure over every configuration of subsets on data points. Using data's similarity matrix and a determinant operator, DPP assigns higher probabilities to those subsets with dissimilar items \cite{Alex2012}. It coincides with the phenomenon which naturally arises in physics (fermions, eigenvalues of random matrices) and in combinatorics (non-intersecting paths, random spanning trees) \cite{HoughBen2006} and is used to capture the repulsion \cite{Sihem2014} among particles. One advantage of DPP is that the inference of DPP can be solved in polynomial time, which is required by many real-time large-scale applications.
We focus on the prior modeling power of DPP and here only name a few recent developments of DPP for interested readers. For DPP (both discrete and continuous models \cite{RajaHafiz20132}), basic inference algorithms, e.g. marginal inference, conditional inference and sampling \cite{Alex2012}, Maximize A Posterior (MAP,  \cite{JenniferGillenwater2012}), are well-developed. Parameter learning of DPP kernels have been recently addressed by \cite{Raja2014} and \cite{gillenwater2014nips}. The technique on measuring the repulsion of DPP was surfaced recently \cite{Christophe2014}.

James Y. Zou \cite{JamesZou2012} 
is the first literature that introduced the DPP prior into generative latent Dirichlet allocation (LDA) model. The intuition is that co-occurring words only appear in a small number of topics. Using a positive definite kernel function, it specifies a preference for diversity over the joint word-topic distributions.

In addition to the DPP prior for LDA \cite{JamesZou2012}, three more DPP-prior based Bayesian models have been developed. \cite{Amar2013} developed a Determinantal Clustering Process (DCP) by using the DPP to define probability measure over point set. Since the diversity prior is placed over all possible partitions of the data, DCP is a nonparametric Bayesian approach, which is useful for semi-supervised clustering task.

Applying DPP as spike-and-slab prior is useful in the context of variable selection under the Bayesian framework. By making use of the repulsion property of DPP, \cite{Mutsuki2014} improved the prediction accuracy of linear regression through the collinear predictors to be less likely selected simultaneously.

J. Snoek and R. P. Adams \cite{Jasper2013} tried to model temporal sequences of spikes to reveal the complexities underlying a series of recorded action potentials. In this paper, DPP models the hidden sequential neurons to capture and  visualize the complex inhibitory and competitive relationships.

\section{Diversified Hidden Markov Models}
\label{sec:dHMM}
In this section, our proposed dHMM and its MAP solution are presented in details.
The graphical model of the proposed dHMM is illustrated in Fig. (\ref{fig:HMMDPP}). In order to make this paper self-contained, we illustrate all of its steps as well as the background knowledge leading up to our new work: (1) We first briefly review the basic models: DPP and HMM. (2) Then, the probability product kernel is introduced as a basic building block for DPP. (3) Our dHMM is subsequently represented. (4) Finally, we detail the inference steps to solve the proposed dHMM.

\subsection{Continuous determinantal point processes}
Determinantal Point Process (DPP) is one popular approach to assign Probability Mass Function (PMF) to each subset of an arbitrary dataset, in either discrete sets space \cite{DPP12}  or continuous sets space \cite{RajaNIPS2013}\cite{RajaNIPS2014}.
By defining a pairwise similarity between the data elements, usually in terms of a Kernel function, DPP assigns higher probability to dissimilar subsets in terms of the determinant of the kernel matrix restricted to the selected subsets. Equally, DPP prefers diverse subsets.

Formally, given a base set $\Omega \subset \mathbb{R}^d$ in a continuous space,
and a positive semidefinite kernel function $K: \Omega \times \Omega \mapsto \mathbb{R}$,
\begin{equation}
\begin{aligned}
P_K(Y) =\frac{det(K_Y)}{\prod_{n=1}^{\infty}(\lambda_n+1)} \nonumber
\end{aligned}
\end{equation}
where $K_Y=[K(x,y)]_{x,y\in Y}$ is the $|Y|\times|Y|$ sub-matrix of $K$ with the restriction to the entries indexed by the elements in $Y$. $\lambda_1, \lambda_2,...$ are eigenvalues of the kernel $K$.
Let $\phi$ be the mapping function: $\phi(x)$, the kernel function can be explicitly expressed as:
\begin{align}
K(x,y) &= <\phi(x),\phi(y)> \nonumber
\end{align}
where $<\cdot,\cdot>$ is the inner product in Hilbert space.
The DPP is geometrically denoted as:
\begin{align}
P_K(Y) &\propto det(K_Y) = vol^2(\{ \phi(x_i)_{ i \in Y }\}) \nonumber
\end{align}
It can be seen that the probability defined by DPP relates to the squared $|Y|$-dimensional volume of the parallelepiped spanned by the selected items in the associated Hilbert space of $K$.
It prefers diverse subsets, because their feature vectors in the Hilbert space are more orthogonal, and hence span larger volumes.

When the cardinality of a diverse subset is fixed to k, which is required by many applications, it is referred as $k$-DPP \cite{kDPPICML2011}. The corresponding probability density $P_K^k$ is given by:
\begin{align}
P_K^k(Y) = \frac{\det(K_Y)}{e_k(\lambda_{1:\infty})}
\label{eq:kdpp}
\end{align}
where $\lambda_{1:\infty}=(\lambda_1, \lambda_2, ...)$ and $e_k(\lambda_{1:\infty})$ is the $k$th elementary symmetric polynomial \cite{kDPPICML2011}.

DPP has closed-form solutions for normalization and marginalization, which makes inference efficient. Dual representation and other efficient approximation methods for large-scale problems have been also developed. Interested readers should refer to \cite{sDPPNIPS2010}\cite{MDPP2014}\cite{MCMCDPP2013}\cite{DPPLVM2013}\cite{DPPLDA2012}\cite{DPPMAP2012} for more details.

\subsection{Kernels for probability diversity}
We prepare a probability kernel, which allows us to apply DPP on HMM's transition probabilities. In this work, the Probability Product Kernel, which is proposed in \cite{ProbabilityProductKernels04},  defines the kernel function between distributions $P_i$ and $P_j$ of discrete variables, which are parameterized by $A_i, A_j$ respectively:
\begin{align}
K(A_i,A_j;\rho)&= <P(x|A_i)^{\rho}, P(x|A_j)^{\rho}>  \nonumber \\
&= \sum_{x \in X} P(x|A_i)^{\rho}P(x|A_j)^{\rho} \nonumber
\end{align}
for $\rho>0$ and $x$ runs through all possible values of discrete variable $X$. The kernel is computed by summing up the products between the two distributions in terms of $x$.

For distributions $P(x|A_i)$ and $P(x|A_j)$, the less `correlation' between them, the more `diversity' we can gain through the determinant of the kernels. To remove the scale effects of different probabilistic measurements, the normalized correlation kernel function is applied:
\begin{align}
\tilde{K}(A_i,A_j;\rho)=\frac{K(A_i,A_j;\rho)}{\sqrt{K(A_i,A_i;\rho)}\sqrt{K(A_j,A_j;\rho)}}
\label{eq:norKernelF}
\end{align}

The final continuous DPP kernel as a building block in our proposed model is:
$\tilde{K}_A = [\tilde{K}(A_i,A_j) ]_{i,j\in \{1,2,...,d\}}$,
where $\tilde{K}_A$ is  $d\times d$ matrix, and  $A_{i\cdot} \in {\mathbb{R}^d_{+}}$ with $\sum_j A_{ij}=1$ for probability measure.

\subsection{Log-likelihood function of Hidden Markov Models }
Hidden Markov models assume what are being observed are generated by a Markov process with unobserved hidden states. It is especially known for their applications in temporal sequential pattern recognition.

In Fig. (\ref{fig:HMMDPP}), a hidden Markov model is a $k$-state Markov chain observed at discrete time points $t = 1,2,...,T$. Let $\{A_1,A_2,..., A_k\}$ be the finite state space. One state $A_i$ can be transfered to all other states $\{A_j, j\in \{1,2,...k\}\}$ with probability distributions parameterized by transition matrix $A \equiv [a_{ij}], i,j \in \{1,2,...,k\}$. We use $X=\{X_1,X_2,...,X_T\}$ as state variables, and $X_t=A_i$ means HMM is staying on state $A_i$ at time step $t$. $P(X_t=A_j|X_{t-1}=A_i)$ denotes the transition probability from $A_i$ to $A_j$, which is equal to $A_{ij}$.  In unsupervised setting, hidden variables cannot be observed directly, which are represented by hollow circles in Fig. (\ref{fig:HMMDPP}). In contrast, filled circles denote observations $Y=\{Y_1,Y_2,...,Y_T\}$. Each chain observation is parameterized by i.i.d emission distribution $B$ given hidden states.
For each hidden state, its probability distribution is only dependent on its former state in the first-order HMM.
The joint probability distribution over hidden variables and observations is parameterized by  $\lambda=(\pi,A,B)$  \cite{ClassicalHMM1989}. The likelihood is as follows.
\begingroup
\allowdisplaybreaks
\begin{align}
P(X_1&,...,X_T, Y_1,...,Y_T|\lambda) = P(X_1;\pi)\prod_{t=2}^{T} P(X_t|X_{t-1};A) \nonumber \\
&~~~~~~~~\times  \prod_{t=1}^{T}P(Y_t|X_t;B) \nonumber \\
s.t. \;\;& \sum_{i=1}^k \pi_i = 1, \pi_i\geq 0, i\in \{1,2,...,k\} \nonumber \\
& \sum_{j=1}^k A_{ij} = 1, A_{ij}\geq 0, \;\; i,j\in\{1,2,...,k\} \nonumber
\label{eq:jointDensity} \\
&B :\; probability\; measure \nonumber
\end{align}
\endgroup
The linear constraints are required by discrete probability measure. The last statement above means that parameters $B$ of emission distributions should also satisfy the requirement of probability measure in either continuous or discrete space, decided by various applications.


Since supervised HMM is a special case of unsupervised HMM  with known hidden state for training period and known parameters for test period, here we just demonstrate the solutions for unsupervised case.
Three basic problems of HMM are identified in \cite{ClassicalHMM1989}, namely, adjusting parameters given observations $max_{\lambda} P(Y|\lambda)$ for unsupervised setting (or $max_{\lambda} P(Y,X|\lambda)$ for supervised setting), computing log likelihood $log P(Y|\lambda) $ for unsupervised setting and inferring hidden states $max_{X}P(Y,X|\lambda)$ given both parameters and observations for both unsupervised setting and supervised setting during test period.

For unsupervised learning, these three problems are closely associated with the likelihood, which is computed  by marginalizing out the hidden variables from the joint distribution. The log likelihood for one sequential observation is:
\begin{equation}
L(Y;\lambda)=log P(Y|\lambda)=log \sum_X P(X,Y|\lambda)
\label{eq:Likelihood}
\end{equation}

With Markov assumption of HMM and Jensen's inequality, the lower bound of the intractable formula in Eq.(\ref{eq:Likelihood}) is:
\begin{align}
L(Y;&\lambda) \geq  \sum_{X_1} q(X_1) logP(X_1|\pi) \nonumber \\
+ &\sum_{i=1}^T \sum_{X_i} q(X_i) log P(Y_i|X_i,B) \nonumber \\
 +& \sum_{i=2}^T \sum_{X_{i-1}, X_i}q(X_{i-1},X_i)logP(X_i|X_{i-1},A) \nonumber \\
-& \sum_X q(X)log q(X) \label{eq:expectation}
\end{align}
where $\{q(X_i)\}_{i=1}^T$ and $\{q(X_{i-1},X_i)\}_{i=2}^T$ are marginally unary and pairwise distributions of hidden variables.

Traditional HMM is solved under EM framework. As noted, it usually produces flatten emission distributions and meaningless transition matrix. In the next subsection, we detail how to encode the diversity-encouraging prior into the transition distributions and how to solve the three basic problems identified by the traditional HMM.

\subsection{Proposed Diversified HMM}
With all the previous concepts in hand, we can now proceed with our proposed diversified HMM (dHMM).

The HMM's  transition distributions $P(X_t|X_{t-1})$ obey multinomial distribution parameterized by $\{A_{ij}\}_{i,j=1}^k$, where $t$ is the time index and $k$ is the number of hidden states. The corresponding normalized correlation kernel function for rows of $A$ based on Eq.(\ref{eq:norKernelF}) is as:
\begin{align}
\tilde{K}(A_{i\cdot},A_{j\cdot}) &= \frac{\sum_{x=1}^k (A_{ix}A_{jx})^\rho}{\sqrt{\sum_{x=1}^k A_{ix}^{2\rho}} \sqrt{\sum_{x=1}^k A_{jx}^{2\rho}}}
\label{eq:prior}
\end{align}
And the corresponding diversity prior of transition parameter matrix of HMM modeled by Determinantal Point Processes (DPP) based on Eq.(\ref{eq:kdpp}) is:
\begin{equation}
P_{\tilde{K}}^k(A) \propto \det(\tilde{K}_{A})
\label{eq:diversityMeasure}
\end{equation}
where $\tilde{K}_{A}$ is $|A| \times |A|$ kernel matrix, $A_i \in \mathbb{R}^k_+$ with $\sum_j A_{ij}=1$.
$A$ is a $k$-size subset from the $k-1$ simplex. $P_{\tilde{K}}^k$ symbols  $k$DPP.
For all experiments based on our proposed dHMM, we set $\rho=0.5$.

\begin{figure*}[!ht]
\center
\includegraphics[width=1.1\textwidth]{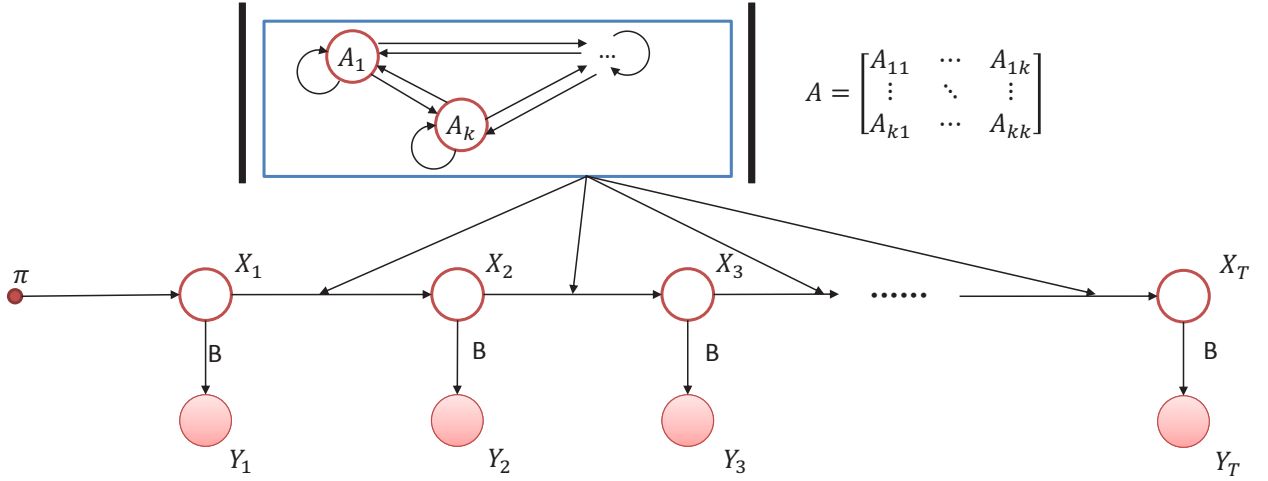}
\caption{Graphical model of diversified HMM }
\label{fig:HMMDPP}
\end{figure*}

The graphical model of our proposed model is illustrated  in Fig. (\ref{fig:HMMDPP}). The bottom chain structure is a standard first-order HMM. Applying conventional symbols of graphical models, hollow circles indicate hidden states, while filled circles are symbols of observations. Similar to \cite{JamesZou2012}, we draw a double-struck plate to denote the DPP prior placed on the state transition matrix $A$. 
Higher the probability of DPP is, the more diverse of the rows of an HMM's transition matrix is.

\vspace{-0.5em}
\subsubsection{Unsupervised setting}
To model the unsupervised sequential labeling,
Maximum A Posterior (MAP) problem needs to be solved since it incorporates diversity-encouraging prior over parameters of rows the transition matrix. The new objective function is formulated as:
\begin{equation}
\begin{aligned}
max_{\lambda} \;\; & L(Y;\lambda) + \alpha log|\tilde{K}_{A}| \\
s.t. &\sum_{i=1}^k \pi_i = 1 , \pi_i \geq 0, i\in \{1,2,...k\} \\
& \sum_{j=1}^k A_{ij}=1, \;\; A_{ij}\geq 0,\;\;i,j\in \{1,2,...,k\}  \\
&B: ~probability ~ measure
\end{aligned}
\label{eq:Optimal}
\end{equation}
where $\lambda=(\pi,A,B)$ is the parameters of our proposed dHMM and we adopt the same symbols with the traditional HMM.
Note that we ignore the normalization constant of the DPP prior distribution, since it is irrelated to measuring the diversity of parameters of rows of transition matrix, as well as estimating parameters of initial distribution and emission distributions.
And $\alpha>0$ is used to balance the weights between measurements of likelihood and diversity-encouraging prior.
When $\alpha = 0$, no diversity-encouraging prior will distract the estimation of transition matrix from Maximum Likelihood Estimation (MLE) learning. With $\alpha$ goes up, the weight of diversity-encouraging prior increases, and the diversity-encouraging prior will dominate the estimation of the parameters of transition matrix.

\subsubsection{Supervised setting}
For modeling supervised sequential labeling, as the hidden states are given during training period, parameters $\lambda=(\pi,A,B)$ can be learned in a count manner. Specifically, $\pi = A_i$ is the ratio between the frequency of state $A_i$ and the total number of sequences. $A_{ij}$ is the proportion of the pairwise states $(X_{t-1} = A_i,X_t=A_j)$ among all pairwise states appearing in the training sequences. $B$ can be learned in a discriminative manner, since the observations are independent given hidden states.
Obviously, the learned parameters  fit the training dataset best, rather than the test dataset. To generalize the counting-computed parameters of transition matrix $A_0$ by incorporating diversity-encouraging prior, we construct the new objective function as below:
\begin{equation}
\begin{aligned}
max_{\lambda} \;\; & L(Y,X;\lambda) + \alpha log|\tilde{K}_{A}| - \alpha_A ||A-A_0||^2_2\\
s.t. &\sum_{i=1}^k \pi_i = 1 , \pi_i \geq 0, i\in \{1,2,...k\}  \\
& \sum_{j=1}^k A_{ij}=1, \;\; A_{ij}\geq 0,\;\;i,j\in \{1,2,...,k\}  \\
&B: ~probability ~ measure
\end{aligned}
\label{eq:supervisedHMMDPP}
\end{equation}
where $A_0$ is the trained parameters by $\lambda_0 = max_{\lambda} L(Y,X;\lambda)$ with $\lambda_0 = (\pi_0, A_0, B_0)$, $\alpha_A$ is used to control how far the final $A$ can drift from $A_0$.

\subsection{Solutions} \label{sec:solutions}
In this subsection, we mainly focus on how to learn parameters from unsupervised setting with objective function Eq.(\ref{eq:Optimal}) and supervised setting with objective function Eq.(\ref{eq:supervisedHMMDPP}).

\subsubsection{Unsupervised setting}
Traditionally, Expectation-Maximization (EM) framework \cite{fang2014active} is applied to learn HMM parameters. Here, our procedure is only different with traditional EM in M-step.
This is because in a MAP setting, the diversity-encouraging prior term, i.e., $\log(\lambda)$ is irrelated to the hidden states $X$, which can be taken out of the integration in E-step.

\textbf{E-step:}

Given old parameters $\lambda^{old} = (A^{old}, B^{old}, \pi^{old})$, we apply forward-backward algorithm to do inference for hidden variables.

In the forward pass of HMM chain,  it inductively summarizes all information before time step $t$ into marginal distribution over each hidden variable $X_t$ and all past observation variables $\{Y_1,...,Y_t\}$, namely,
\begin{align}
\alpha(X_t) &\propto P(X_t,Y_1,Y_2,...,Y_t;\lambda^{old}) \nonumber \\
\alpha(X_{t+1}) &\propto \left( \sum_{X_{t}} \alpha(X_t)P(X_{t+1}|X_t) \right)\times P(Y_{t+1}|X_{t+1}) 
\end{align}

Similarly, in the backward pass, it summaries information over all future observation variables after time step $t$, $\{Y_{t+1},...,Y_{T}\}$, namely,
\begin{align}
\beta(X_t) &\propto P(Y_{t+1},...,Y_T|X_t;\lambda^{old}) \nonumber \\
\beta(X_{t-1}) &\propto \sum_{X_{t}} \left( \beta(X_t)P(X_t|X_{t-1})P(Y_t|X_t) \right) 
\end{align}

The initializations for both forward and backward pass are:
\begin{align}
\alpha(X_1) &\propto P(X_1|\pi^{old})\times P(Y_1|X_1,B^{old}) \nonumber \\
\beta(X_T) &= 1 \nonumber
\end{align}

The conditionally marginal probabilities for hidden variables (required by likelihood in Eq.(\ref{eq:expectation})) and likelihood can be computed by combining the forward and backward summarizations. The unary and pairwise hidden states distributions as well as the normalization are formulated as:
\begin{align}
q(X_t) &\propto \alpha(X_t) \beta(X_t) \nonumber \\
q(X_{t-1}, X_t) & \propto \alpha(X_{t-1}) P(X_t|X_{t-1})P(Y_t|X_t)\beta(X_t) \nonumber \\
P(Y_1,Y_2,...,Y_T) &= \sum_{X_t} \alpha(X_t)\beta(X_t) \nonumber
\end{align}

\textbf{M-step:}
In this step, dHMM optimizes the below objective function to update the parameters $\lambda^{old}=(\pi^{old},A^{old},B^{old})$ given $N$ training sequences.

\begin{align}
&max_{\pi,A,B} L(\pi, A, B|Y,X)  \nonumber \\
=&\sum_{n=1}^N \left( \sum_{t=2}^{T_n}\sum_{X_{nt},X_{n,t-1}}q(X_{nt},X_{n,t-1})logP(X_{nt}|X_{n,t-1},A) \right) \nonumber \\
 +& \sum_{n=1}^N\left( \sum_{t=1}^{T_n}\sum_{X_{nt}}q(X_{nt})logP(Y_{nt}|X_{nt},B) \right)  \nonumber\\
+& \sum_{n=1}^N \left(\sum_{X_{n1}}q(X_{n1})logP(X_{n1}|\pi) \right)
+\alpha log(|\tilde{K}_{A}|) \nonumber \\
&s.t.    \sum_{i=1}^k \pi_i = 1, \pi_i \geq 0, i\in \{1,2,...,k\}  \nonumber \\
&\;\;\;\; \sum_{j=1}^k A_{ij}=1, \;\; A_{ij} \geq 0, i,j\in \{1,2,...,k\} \nonumber  \\
&\;\;\;\; B: \;probability \; measure \nonumber
\end{align}
where, $X_{nt},Y_{nt}$ denote hidden state and observation of the $n$th sequence at time step $t$ respectively, $T_{n}$ denotes the length of the $n$-th sample sequence.
Note that, the last term of Eq.(\ref{eq:expectation}), which is the entropy of marginal  distribution $q(X)$,  is irrelated to model parameters $\lambda$, is simply ignored in M-step.

As both $\log$ function and $\log \det$ function are concave, we directly apply the Lagrange multipliers method to solve the maximization problem in Eq.(\ref{eq:Optimal}). The Lagrange function is given below:
\begin{align}
\Lambda(\pi,A,B,\beta) =& L(\pi,A,B|Y,X) \nonumber \\
-&\beta_0 (\sum_{i=1}^k \pi_i - 1) - \sum_{i=1}^k \beta_i (\sum_{j=1}^k A_{ij}-1) \nonumber
\end{align}
where $\beta_i, i\in \{0,1,...,k\}$ is the Lagrange multipliers.

As the gradients for both $\pi$ and $B$ are the same with traditional HMM, we just list the results. For $\pi$,

\textbf{$\pi_i, i\in \{1,2,...,k\}$}
\begin{align}
\pi_i = \frac{\sum_{n=1}^N q(X_{n1}=i)}{N} \nonumber
\end{align}

For emission distribution, in our experiments, it obeys either Gaussian distribution or multinomial distribution.
For Gaussian distribution,  $Y_t|X_t=A_i \sim \mathcal{N}(B.\mu_i,B.\sigma_i), i \in \{1,2,...,k\}$, the updating parameters for $B$ are:

$B.\mu_i$
\begin{align}
\mu_i = \frac{\sum_{n=1}^N \sum_{t=1}^{T_n}Y_{nt}q(X_{nt}=A_i)}{\sum_{n=1}^N \sum_{t=1}^{T_n}q(X_{nt}=A_i)}
\end{align}

$B.\sigma_i^2$
\begin{align}
\sigma_i^2 = \frac{\sum_{n=1}^N \sum_{t=1}^{T_n} q(X_{nt}=A_i)(Y_{nt}-\mu_i)^2}{\sum_{n=1}^N \sum_{t=1}^{T_n}q(X_{nt}=A_i)} 
\end{align}

For multinomial distribution, $Y_t|X_t=A_i \sim \mathcal{M}ulti(B.b_i), i \in \{1,2,...,k\}$,
and the updating parameters for $B$ are:

$B.b_i$
\begin{align}
b_{ij} = \frac{\sum_{n=1}^N \sum_{t=1}^{T_n} q(X_{nt}=A_i) \delta(Y_{nt}=j)}{\sum_{j=1}^k \sum_{n=1}^N \sum_{t=1}^{T_n} q(X_{nt}=A_i) \delta(Y_{nt}=j)} \nonumber
\end{align}
where $\delta(\cdot)$ is the Dirac delta function. The same results are applied when the emission distributions obey the Bayes Bernoulli distribution since Bernoulli distribution is a special case of multinomial distribution. Namely, $Y_t=y_j|X_t=A_i \propto b_{ij}^{y_j} (1-b_{ij}^{1-y_j})$ with $y_j \in \{0,1\}$ and parameters $0\leq b_{ij} \leq 1$.

When handling with transition matrix $A$, the situations for traditional HMM and dHMM are different.
Let $\Lambda_A$ be the Lagrange function related to parameters $A$, namely,
\begin{equation}
\begin{aligned}
&\Lambda_A=\\
& \sum_{n=1}^N \sum_{t=2}^{T_n} \sum_{X_{n,t-1},X_{n,t}} \left( q(X_{n,t-1},X_{n,t}) log(P(X_{n,t}|X_{n,t-1},A)) \right)  \\
&~~~+ \alpha log |\tilde{K}_{A}| - \sum_{i=1}^k \beta_i (\sum_{j=1}^k A_{ij}-1)
\end{aligned}
\label{eq:LangrangeA}
\end{equation}

The gradients corresponding to parameters $A$ are computed by
\begin{equation}
\nabla_{A_{ij}} {\Lambda}_A  = 0
\label{eq:gradientA}
\end{equation}

When $\alpha=0$, the updates for $A$ are the same with traditional HMM:
\begin{align}
A_{ij} = \frac{\sum_{n=1}^N \sum_{t=2}^{T_n} q(X_{n,t-1}=i,X_{nt}=j)}{\sum_{n=1}^N \sum_{t=2}^{T_2} \sum_j q(X_{n,t-1}=i,X_{nt}=j)} \nonumber
\end{align}

When $\alpha>0$, the solution for Eq.(\ref{eq:gradientA}) has no closed form. We iteratively maximize Eq.(\ref{eq:LangrangeA}) with projected gradient ascend method.
First the gradients are computed by:
\begin{equation}
\partial L_{A_{ij}} = \sum_{n=1}^N \sum_{t=2}^{T_n}\frac{q(X_{n,t-1},X_{nt})}{A_{ij}}+\alpha \nabla_{A_{ij}} log|\tilde{K}_{A}|
\end{equation}
with  $\nabla_{A_{ij}} log|\tilde{K}_{A}| = \frac{1}{2}\sum_{m=1}^k \left( [\tilde{K}_{A}^{-1}]_{mi} \frac{\sqrt(A_{mj})}{\sqrt(A_{ij})}  \right)$.
The updates for $A$ are
\begin{equation}
A^{new} = A^{old}+\gamma \cdot \partial L_{A}
\end{equation}
where $\gamma$ is the step size, and here we apply adaptive step in our implementation.

Then we project all rows of $A$ onto the $k-1$ probability simplex by finding a nearest point in the simplex for $A^{new}$, equally, we try to solve the following optimization problem:
\begin{equation}
\begin{aligned}
min_{a_i} &||a_i-A_{i\cdot}^{new}||^2 ~~i = 1,...,k\\
s.t. & a_i^T \textbf{1} = 1,~~a_i \geq 0 \\
\end{aligned}
\label{eq:projectSimplex}
\end{equation}
We refer readers to Algorithm 1 in \cite{wang2013projection} for more details.

The overall procedure for updating transition parameter matrix is summarized in Algorithm \ref{alg:updateA}.
\begin{algorithm}
\caption{updating $A$ for dHMM}
\begin{algorithmic}[1]
\REQUIRE{Initialization $A^{old}$, initial step $\gamma$, error threshold $\delta$}
\STATE $A^{new}=A^{old}$, ${L}_A^{old}$
\REPEAT
\STATE \textbf{Compute gradient} $\partial{L}_{A}$
\STATE find the suitable step size $\gamma$
\STATE $A^{new} \leftarrow A^{old} +\partial{L}_{A}\times \gamma $
\STATE \textbf{Project onto the probability simplex}:
\STATE $A^{new} \leftarrow ProjSimplex(A^{new})$ ~(Algorithm 1 in \cite{wang2013projection})
\STATE compute $L_A^{new}$ with $A^{new}$
\IF {$|L_A^{new}-L_A^{old}|<\delta$}
\STATE break;
\ENDIF
\STATE $A^{old} \leftarrow A^{new}, L_A^{old} \leftarrow L_A^{new}$
\UNTIL {$|L_A^{new}-L_A^{new}|<\delta$}
\RETURN $A^{new}$
\end{algorithmic}
\label{alg:updateA}
\end{algorithm}

The rows of input $A^{old}$ is initialized by samples from Dirichlet distribution and as shown, our stop criterion is based on the likelihood contributed by parameters $A$. The most time-consuming step is to compute the gradients, which are obtained by matrix inversion operation. Fortunately, we usually maintain a small transition matrix to be manipulated.

\subsubsection{Supervised setting}
To solve the optimization problem in Eq.(\ref{eq:supervisedHMMDPP}), again the projected gradient ascend method is applied. And the gradients with regard to $A_{ij}$ are computed as:
\begin{equation}
\begin{aligned}
\partial L_{A_{ij}} = \sum_{n=1}^N \sum_{t=2}^{T_n} & \frac{\delta(X_{n,t-1}=i,X_{n,t}=j)}{A_{ij}}+\alpha \nabla_{A_{ij}} log|\tilde{K}_{A}| \\
&- 2\alpha_A (A_{ij}-A_0{_{ij}})
\end{aligned}
\label{eq:supervisedLearning}
\end{equation}
From above, the pairwise hidden states are counted rather than inferred in the supervised setting. $ \nabla_{A_{ij}} log|\tilde{K}_{A}|$ is the same as in the unsupervised setting. Again, we iteratively update $A$ with the initialization $A_0$ by gradient ascend method until converged.

Finally, we apply Viterbi algorithm to find the most likely hidden state sequences by solving the problem $max_X P(X,Y|\lambda)$ for unlabeled sequential observations under both unsupervised and supervised settings.

\subsubsection{Convergence analysis}
In the unsupervised setting, we maximize $L(Y;\lambda)+\alpha \log |\tilde{K}_A|$ which is lower bounded by $\mathcal{L}(q,\lambda)+ \alpha \log |\tilde{K}_A|$ \cite{bishop2006pattern}.
The E-step is the same with the general EM algorithm for HMMs. With fixed $\lambda^{old}=(\pi^{old},A^{old},B^{old})$, in E-step, the exact posterior distributions of hidden states are derived by maximizing the lower bound of the likelihood $\mathcal{L}(q,\lambda)$, i.e., $\mathcal{L}(q^{\star},\lambda^{old}) \geq \mathcal{L}(q,\lambda^{old})$, where $q^{\star}$ is the optimal posterior distributions. In M-step, $\mathcal{L}(q^{old},\lambda) + \alpha \log |\tilde{K}_A|$ is maximized by using the gradient ascend algorithm, i.e.,  $\mathcal{L}(q^{old},\lambda^{\star}) + \alpha \log |\tilde{K}_{A^{\star}}| \geq \mathcal{L}(q^{old},\lambda^{old}) + \alpha \log |\tilde{K}_{A^{old}}|$, where $\lambda^{\star}$ corresponds to the local maximum of the objective function. Therefore, we can conclude that the EM optimization produces a sequence of objective value that converges to a local maximum.

Similarly, in the supervised setting, the sequence of the objective value produced by the gradient ascend method will also converge to a local maximum.

\section{Experiments}
\label{sec:exps}
In this section, we demonstrate the effectiveness of our proposed diversified HMM (dHMM) by conducting experiments on both simulated and real-world datasets.

\subsection{Toy experiment}
\begin{figure*} 
\vspace{-1em}
\center
\subcaptionbox{{Transition matrix $A$}\label{fig:toy_A} } 
 {\includegraphics[width=0.52\textwidth]{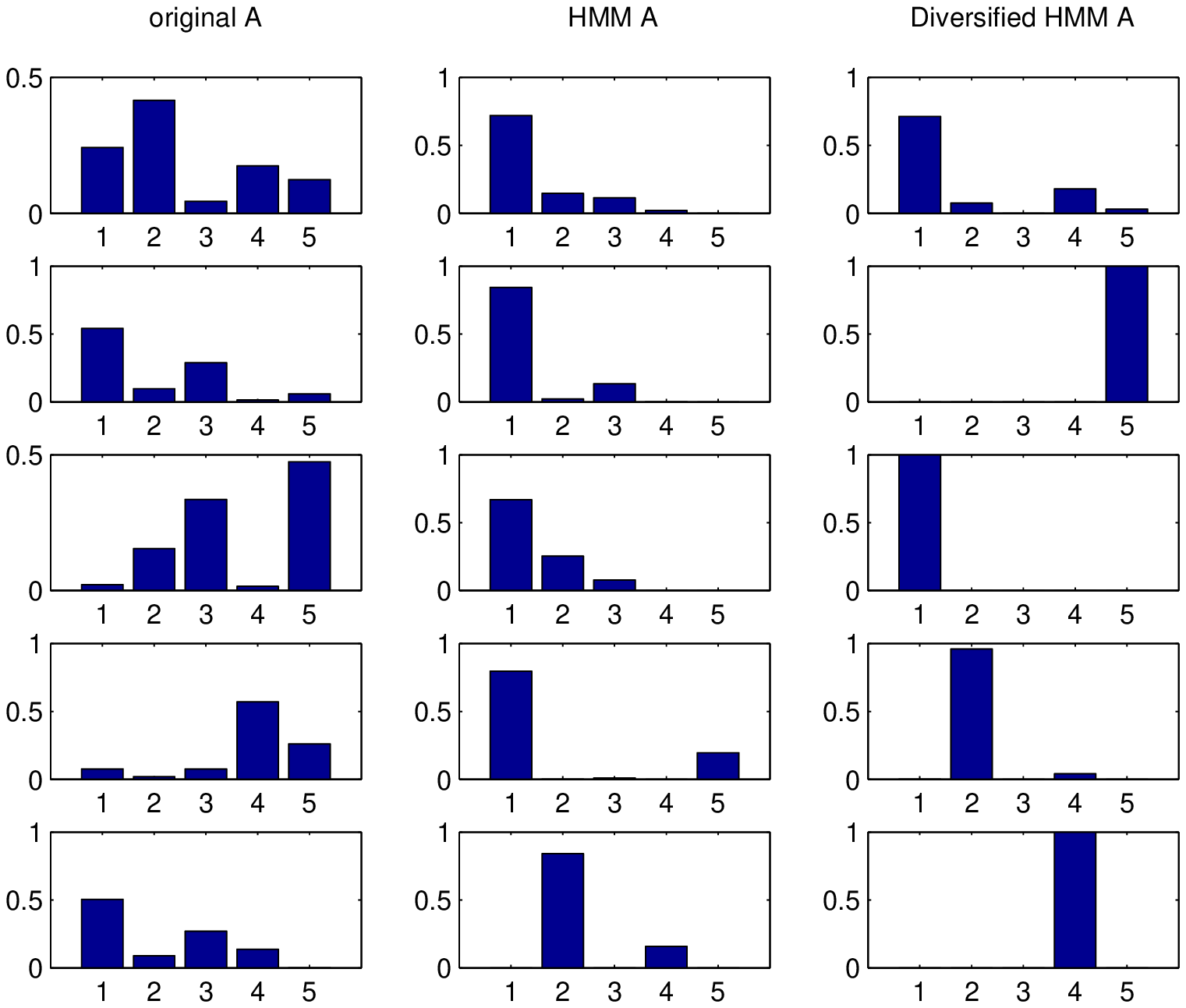}} \hspace{-3em}
\subcaptionbox{{Initial distribution $\pi$ and emission distribution $B$}\label{fig:toy_piB}}
{\includegraphics[width=0.52\textwidth]{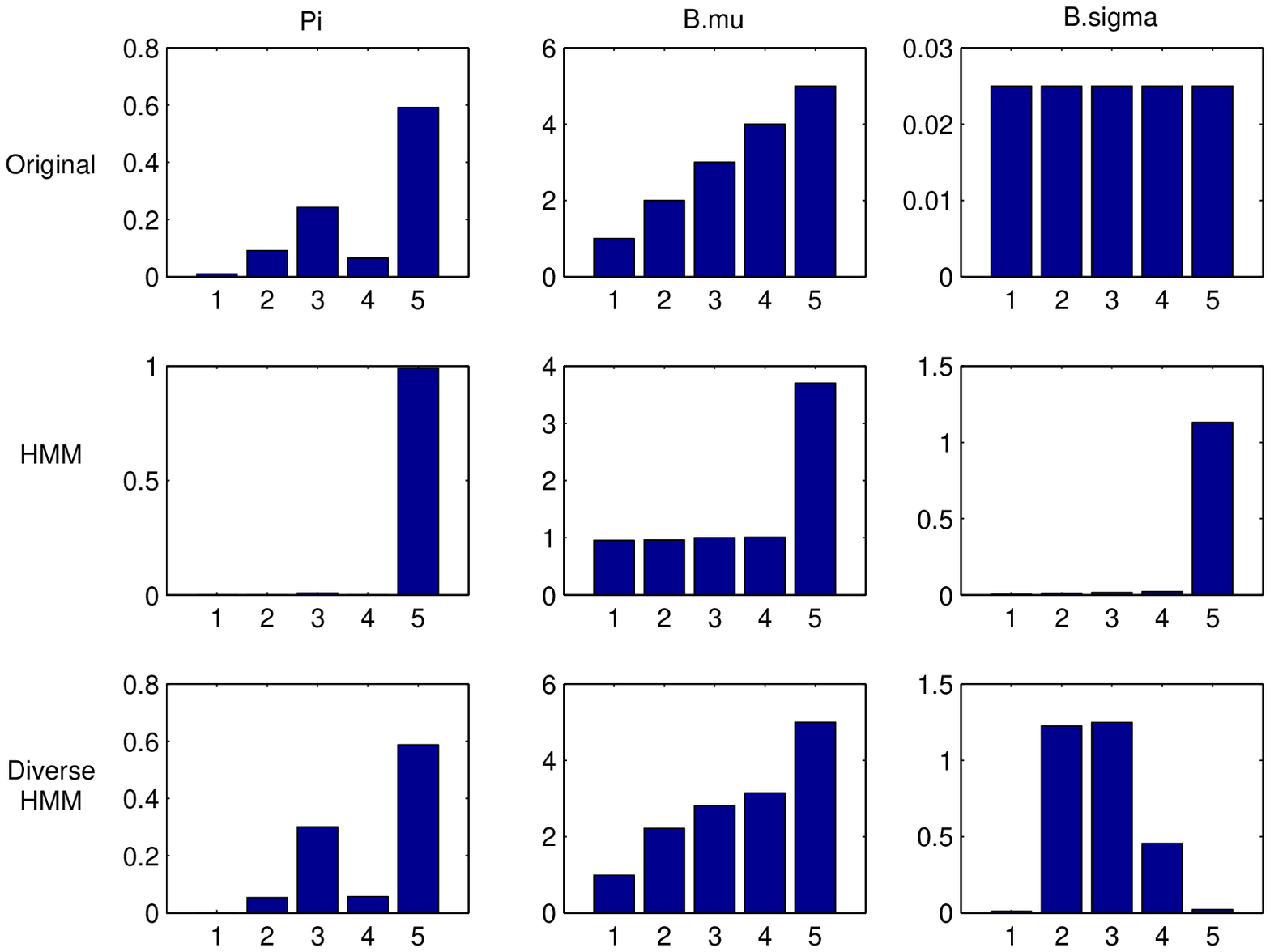}}
\caption{Parameters of ground-truth, learned by proposed dHMM and by traditional HMM}
\label{fig:toyexps}
\end{figure*}

For simulated dataset, $\{1,2,3,4,5\}$ as state space, where the cardinality of the state space is $k=5$. For the ground-truth initial hidden state distribution, it is set as
{
    \def\OldComma{,}
    \catcode`\,=13
    \def,{%
      \ifmmode%
        \OldComma\discretionary{}{}{}%
      \else%
        \OldComma%
      \fi%
    }%
$\pi=(0.0101,0.0912,0.2421,0.0652,0.5914)$. %
}
The transition matrix $A$ is shown in the first column of Fig. (\ref{fig:toy_A}).
The emission probabilities are chosen to be single mode Gaussian distributions. The parameters of means and variances for $k$ Gaussian distributions are set as $B.\mu = (1,2,3,4,5)$, and $B.\sigma_i=0.025, i\in \{1,2,3,4,5\}$.

$300$ observation sequences  were randomly generated from the ground-truth parameters above. For simplicity, we equally set length of all sequences as six, namely, $T_n=6, n \in \{1,2,...,300\}$. The EM framework represented in Sec \ref{sec:solutions} was applied to learn parameters $\lambda=(\pi,A,B)$ for both the traditional HMM and dHMM. The parameters of mean and variance of the emission distributions were initialized with samples from Gaussian distribution and Gamma distribution respectively. Initial state distribution and rows of transition matrix were sampled from a Dirichlet distribution $Dir(\eta_i)$, where the concentration parameters are set as $\eta_i = 3, i\in\{1,2,3,4,5\}$. For diversified HMM, the balance parameter is set as $\alpha=1$. The learned parameters are shown in Fig.(\ref{fig:toyexps}).

\subsubsection{dHMM vs. HMM on Toy dataset}
Since no label information for the learned parameters, alignment between learned parameters and the ground-truth is applied for visualization of the intuitive comparison.
The rows of learned transition matrix  are aligned by minimizing the distance between the learned transition matrix and the ground-truth transition matrix. The learned initial distribution and emission distributions are also aligned with the ground-truth ones accordingly.

In Fig.(\ref{fig:toy_A}), each column corresponds to one $5 \times 5$ transition matrix and each row corresponds to one state's transition distribution. The first column denotes the ground-truth transition distribution, and the last two columns are transition matrices learned from the traditional HMM and diversified HMM respectively.
Comparing to the result of traditional HMM (the middle column), the diversity-encouraging prior takes effect as illustrated in the third column: The transition distributions of different states are mutually distinct. Until now, it is still hard to decide which model infers the hidden states better, since different hidden structures inferred by different models may inherit different meanings.

In Fig.(\ref{fig:toy_piB}), each row illustrates the other two kinds of parameters ($\pi,B.\mu, B.\sigma$) - the first column denotes the state initial distribution while the other two columns denote the means and covariances of the five emission Gaussian distributions. The first row shows the ground-truth. The second row shows the MLE result learned from the traditional HMM. The last row shows the MAP result learned from the proposed dHMM.
From the figure, the traditional HMM identifies only two groups of patterns: The states $1,2,3,4$ are in the first group, which have quite similar emission distributions in terms of their Gaussian means and variances. The state $5$ is in the second group, which has very different Gaussian  parameters comparing to the others. In this case, hidden states $1,2,3,4$ are difficult to be differentiated, which can lead to ambiguous labeling results.
In contrast, our proposed diversified HMM shows superiority in terms of its higher discriminative ability for differentiating the hidden states involved. This is revealed in the third row of Fig.(\ref{fig:toy_piB}), which shows that different hidden states induce distinct emission components. From the results obtained, the claim that the proposed diversity-encouraging prior over rows of transition matrix indirectly increases the discrimination of hidden states is justified.

We also compared the proposed diversified HMM with traditional HMM in terms of sequential labeling accuracy. Given the learned parameters, the most likely sequential labels are inferred  for each observed sequence by Viterbi algorithm.
Both the inferred state frequencies and labeling accuracies of sequential labeling for different models are summarized in Table (\ref{tab:simulatedAcc}).
The histograms of the sequential labels (i.e., the frequencies of hidden states in the given dataset) inferred from ground-truth parameters, parameters learned from traditional HMM and parameters learned from diversified HMM are shown in the second row of Table (\ref{tab:simulatedAcc}). The ground-truth histogram distributes almost equally amongst the five states, while the statistic of hidden states inferred from traditional HMM tends to be highly biased which favors one dominant state. This problem is somewhat rectified by the proposed diversified HMM. Shown in the third column, the histogram of hidden states inferred from dHMM shows more resemblance to the ground-truth than the histogram inferred from traditional HMM.

To compute the 1-to-1 accuracy of sequential labeling, labels inferred by parameters learned from both the traditional HMM and diversified HMM are aligned to the ground-truth by Hungarian algorithm. As shown in the third row of Table (\ref{tab:simulatedAcc}), the proposed dHMM outperforms traditional HMM by a large margin.

\begin{table*}[t]
\caption{Comparison of state frequencies and accuracies between dHMM and HMM}
\newcommand{\tabincell}[2]{\begin{tabular}{@{}#1@{}}#2\end{tabular}}
\begin{center}
\begin{tabular}{|p{1.3cm}|c|c|c|}
\hline
\tabincell{c}{\\ \\}&ground-truth&HMM&dHMM\\ \hline
&&&\\
\tabincell{c}{state \\ histograms } &\includegraphics[width=0.24\textwidth]{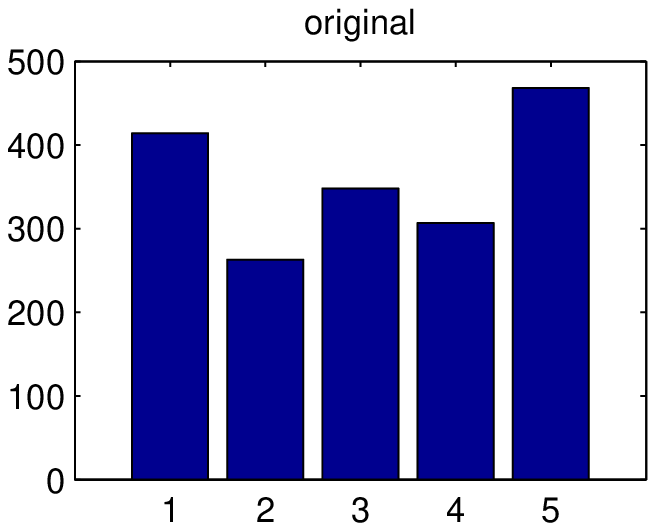}&\includegraphics[width=0.24\textwidth]{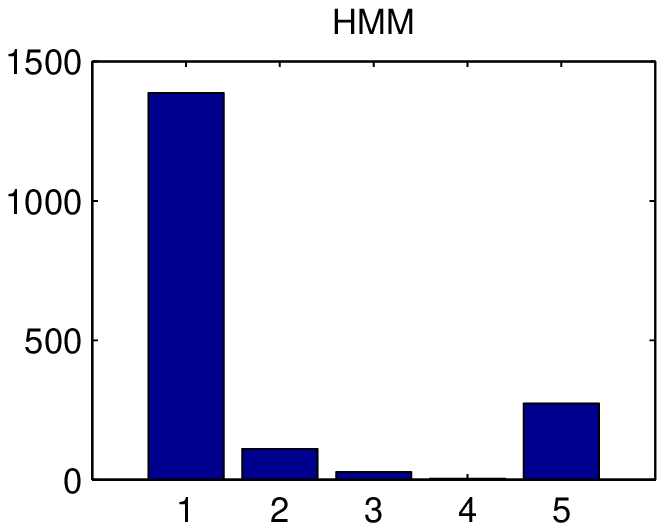}&\includegraphics[width=0.24\textwidth]{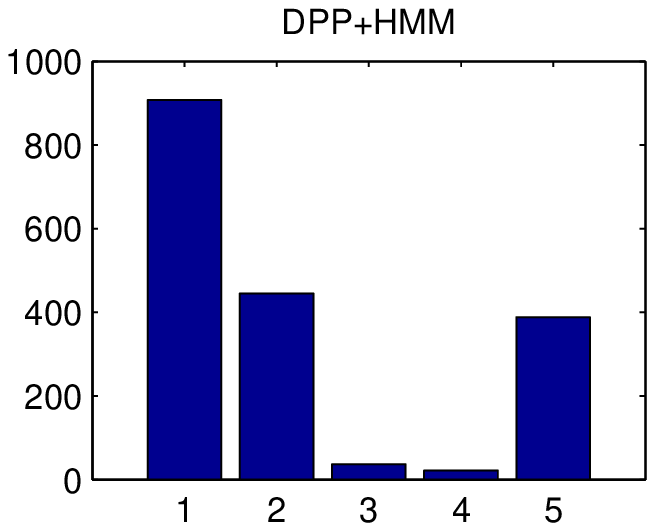} \\
&&&\\
\hline
\tabincell{c}{labeling\\ accuracies}&1&0.4117&\textbf{0.4728} \\ \hline
\end{tabular}
\label{tab:simulatedAcc}
\end{center}
\vspace{-2em}
\end{table*}

\subsubsection{More explanations on dHMM's superiority over HMM}
Further, in this subsection, the superiority of our proposed dHMM over traditional HMM is statistically illustrated especially in the case where the emission distributions are almost flatten. Under this situation, the hidden states are ambiguous and less discriminative, which leads to that traditional HMM identifies less hidden states than the ground-truth, i.e. the learned transition matrix contains more similar rows than the ground-truth transition matrix does. Not only intuitively but also experimentally, our proposed dHMM mitigates this issue to some extent.

All ground-truth parameters of HMM take the same setting as above except the variances of the emission distributions $B.\sigma_i, i\in \{1,2,3,4,5\}$. The variances of the Gaussian distributions are gradually enlarged to `flatten' the emission distributions. In here, we used a sequence of variance parameters $\{B_t.\sigma_i,i\in{1,...,5}\}_{t=1}^T, T=50$, where $B_t.\sigma_i = 0.025+0.1\times(t-1)$. For each $t$, we generated the experimental sequences by the same method described above. The experimental results are averaged over 10 runs with independent initializations.

\begin{figure}
\centering
\includegraphics[width=0.52\textwidth]{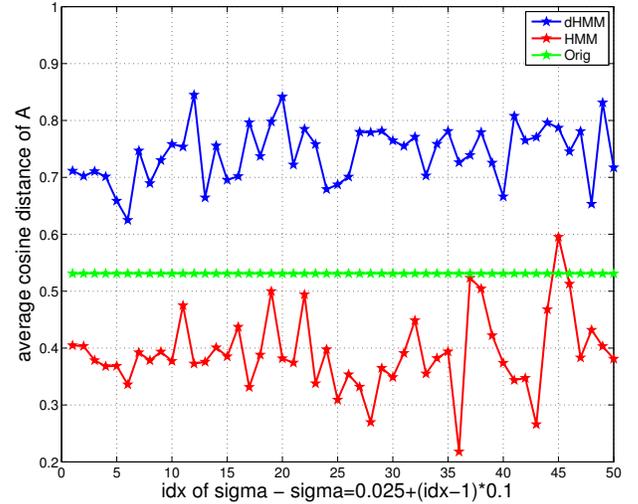}
\caption{Diversities of transition matrix of ground-truth, dHMM-learned and HMM-learned with regard to the parameter of variance of the Gaussian emission distributions}
\label{fig:diverseStatistics}
\end{figure}

We apply averaged Bhattacharyya distance over all pairwise of rows of transition matrix as diversity measure. Higher Bhattacharyya distance means more diversity of rows of transition matrix. The quantized diversities of rows of transition matrix is shown in Fig. (\ref{fig:diverseStatistics}). The green line shows the diversity of the ground-truth transition matrix whose value is $0.531$. The red curve below the green line and the blue curve above the green line show the diversities of the transition matrices learned by traditional HMM and proposed dHMM respectively. The effectiveness of diversity-encouraging prior is obvious: The dHMM consistently outperforms the HMM, no matter what the parameters of variances are set.

\begin{figure}
\centering
\includegraphics[width=0.53\textwidth]{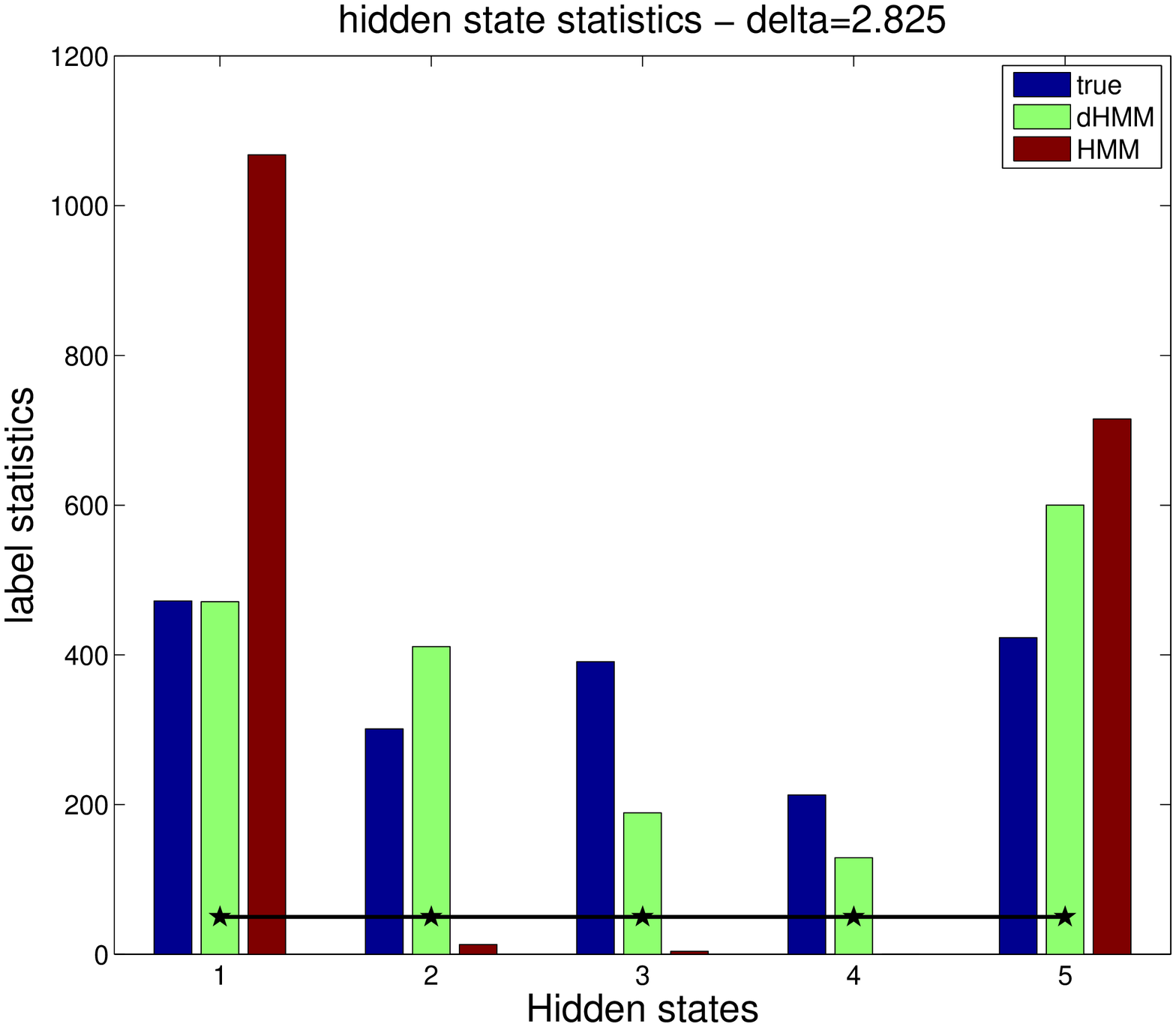}
\caption{Histograms of hidden states inferred from parameters of ground-truth, dHMM-learned and HMM-learned}
\label{fig:stateStatisticsDelta}
\vspace{-2em}
\end{figure}

Higher diversity of transition matrix implies more inferred hidden states.  One example of the histogram of the inferred states is shown in Fig.(\ref{fig:stateStatisticsDelta}), in which the number of states is identified by omitting the labels whose frequencies are below certain threshold $\sigma_F$. In our case, we set the threshold as $\sigma_F=50$ indicated by the black line.
We show that the dHMM (colored as Green) identifies all five states, while the HMM (colored as Red) identifies only two states, since the frequencies of states $2,3,4$ are below the threshold $\sigma_F$.

\begin{figure}[!ht]
\centering
\includegraphics[width=0.53\textwidth]{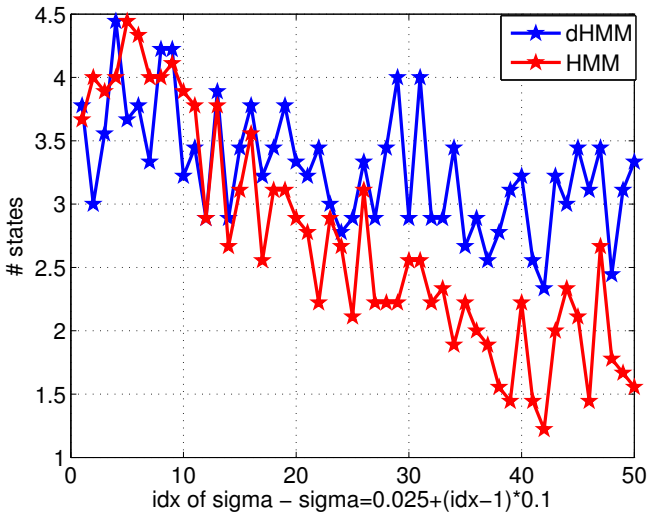}
\caption{Number of hidden states inferred by model parameters of dHMM-learned and HMM-learned with regard to the variance of Gaussian emission distributions}
\label{fig:numStatesStatistics}
\vspace{-2em}
\end{figure}

We summarize the statistical results in Fig.(\ref{fig:numStatesStatistics}). From the left part of the curve, the emission Gaussian distributions start with low variance. The hidden states can be easily identified and the dHMM performs on par with HMM.
However, along with the increasing of the variance, the emission Gaussian distributions are
becoming increasingly `flattened', which make the states severely ambiguous and hard to identify. Shown in the right half part of the curve, the advantage of our dHMM is becoming more obvious as it identifies more hidden states than the traditional HMM does.

\subsection{Real-world experiments}
In this section, our proposed diversified HMM (dHMM) is applied to solve real-world sequential labeling problems: PoS tagging under unsupervised setting and OCR under supervised setting. 

\subsubsection{PoS tagging}
Part-of-Speech tagging (PoS tagging) \cite{CL2003} has been
used in the linguistics community for a long time. The task is to automatically assign contextually appropriate grammatical descriptors to words in texts. In fact, PoS tagging usually produces low level semantic information, which can serve as a precursor towards more abstract levels of analysis, e.g., text indexing and retrieval, as nouns and adjectives are better candidates for index terms than adverbs or pronouns are.

\begin{figure}[t]
\center
\includegraphics[width=0.5\textwidth]{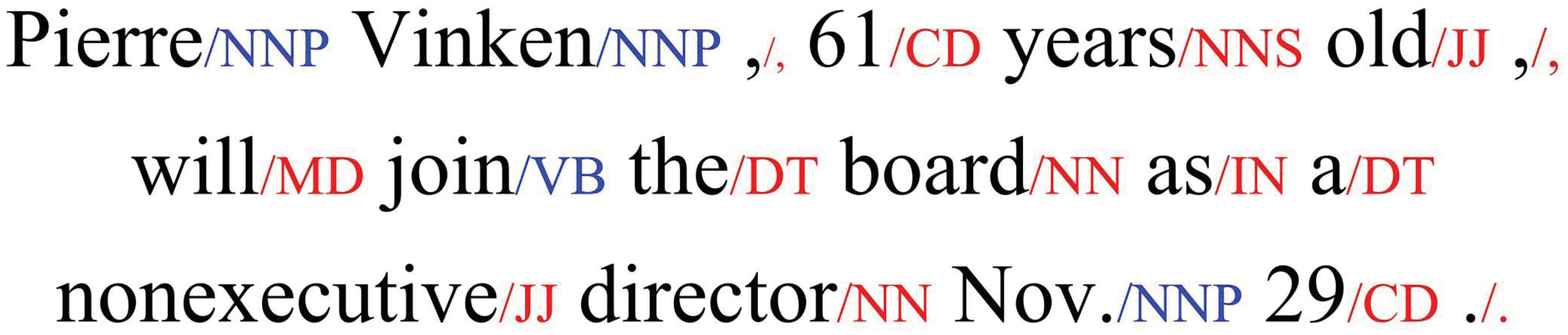}
\caption{Sentence example with PoS tags  }
\label{fig:sentence}
\end{figure}

The Penn Treebank Wall Street Journal (WSJ) corpus (\cite{PennTreebank93}) is one of the most widely used datasets for evaluating performance of the statistical language models.
The training corpus, tagged by gold standard PoS tags, is utilized to evaluate proposed diversified HMM (dHMM) for Part-of-Speech (PoS) tagging task under unsupervised setting.
The vocabulary size of the corpus is around $10K$.
In Table (\ref{tab:POStaggers}), the PoS tags appeared in the WSJ corpus are listed. Detailed definitions of the abbreviation of tags are annotated in \cite{PennTreebank93}.
The tags are preprocessed to reduce the hidden state size from $46$ to $15$ by combining similar tags. The $idx$ column shows the indexes of the reduced tag set. The $PoS$ column shows the semantic title of the tags. The $frequency$ column shows the frequencies of all tags. From this statistics, $25\%$ tags account for nearly $85\%$ words.
All of $3828$ sentences are used in our experiment, and the sequential length is between $2 \sim 250$. An example sentence with true sequential PoS tags is illustrated in Fig. \ref{fig:sentence}, where the true tags lay behind the corresponding words. Naturally, the transition distribution for different tags are different. Take tags /NNP and  /VB as an example, the /NNP has higher probability to be followed or following the same /NNP tag. By contrast, /VB is usually followed by /DT or /IN, and follows /MD, /TO or /RB. This discriminative prior information considered by our model is significantly helpful for sequential labeling, which is verified and demonstrated in detail below.

\begin{table}[t]
\caption{Summary of PoS tags of WJS corpus}
\begin{center}
\begin{tabular}{rcc|ccc}
\hline
idx & PoS &frequency &idx & PoS &frequency  \\ \hline
1.& NNP &9408 &    6.& VBD &3043    		\\ 
1. & NNPS & 244  & 6.& VBN &2134          \\
1.& NNS &6047 &   6. & VBP & 1321          \\ 
1.& NN &13166 &   6. & VBG|NN & 1  	    \\ 
1. & SYM & 1  &  7.& DT &8165   	\\ 
2.& , &4886 &  7. & PDT & 27 	\\
2. & -- & 712 & 7. & WDT & 445    \\ 
2. & '' & 693   &8.& IN &9959  \\ 
2. & : & 563 &  8.& CC &2265    	\\ 
2.& . &3874 & 8. & TO & 2179 \\ 
2. & \$ & 724 & 9. & FW & 4   \\ 
2. & ( & 120 & 10. & WRB & 178  	\\
2. & ) & 126 & 10. & RB & 2829   \\
2. & LS & 13 & 10. & RBS & 35  \\ 
2. & \# & 16 & 10. & RBR & 136   \\ 
3.& CD &3546 & 11. & UH & 3   \\ 
4. & JJS & 182  & 12. & WP & 241 	\\
4.& JJ &5834 &  12. & WP\$ & 14 	\\
4. & JJR & 381 &12. & PRP & 1716  \\
5. & MD &927 & 12. & PRP\$ & 766   \\ 
6.& VBZ &2125 & 13. & POS& 824  	\\ 
6.& VB &2554 &  14. & EX & 88  	\\ 
6.& VBG &1459 &  15. & RP & 107 	\\ \hline
\end{tabular}
\label{tab:POStaggers}
\end{center}
\vspace{-2em}
\end{table}

The number of hidden states is set as $k=15$ as enumerated in Table (\ref{tab:POStaggers}).
Afterwards, the initial state distribution $\pi$ is a $15$-dimension vector and the transition distribution is parameterized by $A$ which is a $15\times 15$ matrix.  The words in the vocabulary are treated as observations and the emission distributions are parameterized by $B$ which is a $k \times V$ matrix, where $V$ is the size of vocabulary. The $\pi$, each row of $A$ and each row of $B$ are randomly initialized by samples from the Dirichlet distribution. We apply the 1-to-1 accuracy measure to quantize our experimental results. Similar with the simulated experiment, the Hungarian algorithm is utilized to map the inferred labels to the ground-truth ones.

\begin{figure}
\center
\includegraphics[width=0.45\textwidth]{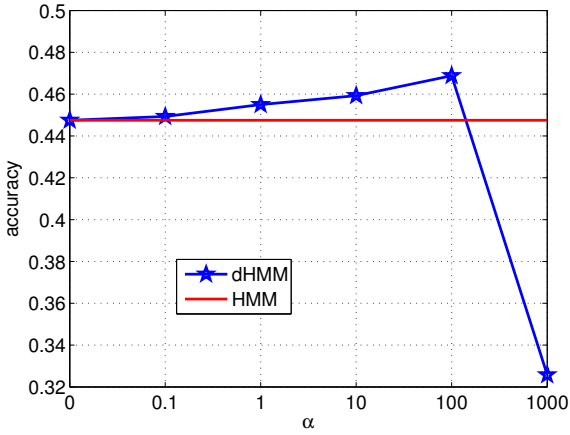}
\caption{Effectiveness of $\alpha$ for PoS tagging}
\label{fig:PoS_alpha}
\vspace{-2em}
\end{figure}

First, we test the effectiveness of diversity-encouraging prior over rows of transition matrix in terms of prior weights, namely, $\alpha$ values. The labeling accuracies with regard to $\alpha$s are plotted in Fig.(\ref{fig:PoS_alpha}). The setting $\alpha=0$ corresponds to the setting of traditional HMM.
Traditional HMM gets an accuracy of $0.4475$, while our proposed dHMM achieves the best accuracy of $0.4688$ with  $\alpha=100$. Larger weight ($\alpha$) will overemphasize the diversity-encouraging prior over rows of transition matrix and will lead to decreasing the sequential labeling accuracy,  as shown with a sharp drop when $\alpha$ increases to $1000$ in Fig.(\ref{fig:PoS_alpha}).

\begin{figure}[t]
\center
\includegraphics[width=0.45\textwidth]{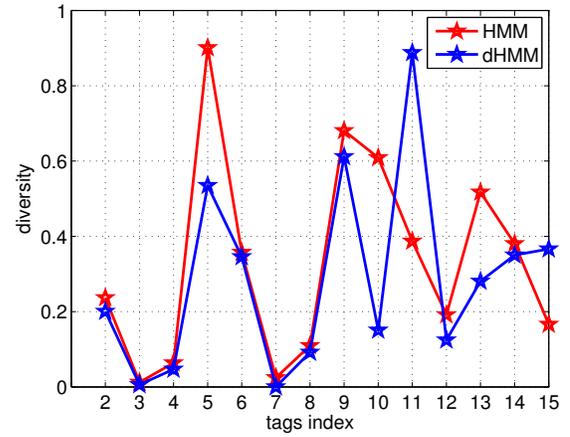}
\caption{Transition diversity comparison between dHMM and HMM for tag `1' and all other tags}
\label{fig:pos_diversity}
\end{figure}

Then, we qualitatively demonstrate the effectiveness of our diversity prior by comparing the transition parameters matrix learned from dHMM ($\alpha=100$) to the parameters learned from the baseline - traditional HMM.
The diversity measurements (Bhattacharyya distance)  between tag $1$ and the other tags are shown in Fig. (\ref{fig:pos_diversity}). The proposed dHMM identifies that tag $1$ (NOUN) is most different from tag $11$ (Interjection), while HMM identifies that tag $1$ (NOUN) is most different from tag $5$ (MODAL).
Since the frequency of tag $11$ (Interjection) is only $3$, intuitively, the transition distribution of this tag is quite different from the transition distributions of other tags. The same situation is applied to tag $9$ (Foreign word, whose frequency is $4$). From Fig. (\ref{fig:pos_diversity}), the proposed dHMM identifies more accuracy result, which indicates that the transition distributions of both tags $11$ and $9$ are most different from tag $1$ (NOUN), than the result learned from the traditional HMM.

\begin{figure}[t]
\center
\includegraphics[width=0.45\textwidth]{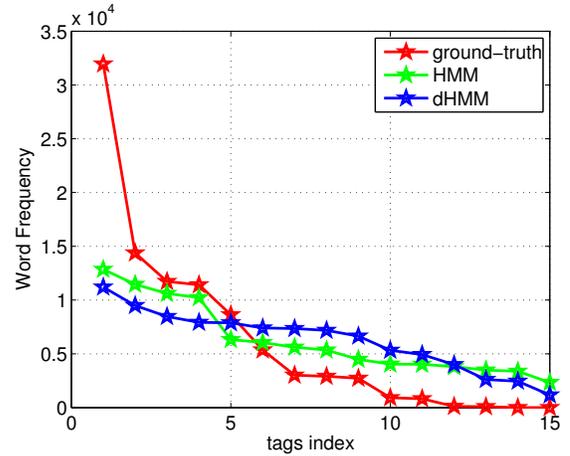}
\caption{Histogram comparisons among ground-truth, HMM and dHMM}
\label{fig:POSemissionCurve}
\vspace{-2em}
\end{figure}

Finally, we show how the diversity-encouraging prior indirectly rectifies the emission distributions learned from traditional HMM to fit the dataset better, as illustrated in Fig. (\ref{fig:POSemissionCurve}). Here, we choose $\alpha=100$ to explain the behavior of the proposed dHMM.
Three curves are plotted. The statistics of the `ground-truth' curve is obtained through the inferred hidden labels by the true parameters.
\footnote{obtained by counting the starting tags, the pairwise tags, and the tag-word pairs of each sentence through the whole corpus, these three statistics are corresponding to physical meanings of the HMM parameters $\lambda=(\pi,A,B)$: initial distribution, transition distribution and emission distributions.}
From the `ground-truth' curve, small portion of tags explain majority of the words, which is pointed out as skewed long-tail distribution \cite{WhynotEM07}. `dHMM' curve learned by our proposed diversified HMM reflects this phenomenon especially for the less frequent $10$ tags and shows promising result for unsupervised sequential labeling task. To some extent, the diversified transition prior latently adjusts the flatten distributions for the less frequent $10$ tags obtained from traditional HMM to a better trend approaching the true distributions.

\subsubsection{OCR}
Optical character recognition (OCR) is a task of converting images of typewritten or printed texts (which are the common forms of scanned data, e.g., passport, receipts) into computer-readable texts. It can be applied to many real-world applications, such as efficient data entry for business documents, automatic number plate recognition.
To achieve this aim, one has to perform both character segmentation and recognition tasks. In this work, we assume every character has been segmented out and stored in its own image patch, so that we focus on the recognition task, which is referred as sequential labeling here.

We apply the OCR dataset processed by \cite{MMMN2003}. They select clean subset from the handwritten words collected by Rob Kassel at the MIT Spoken Language Systems Group. By removing the first capitalized letters, Ben et al. rasterized and normalized images of the rest lowercase letters into $16 \times 8$ images. There are in total 6877 words containing $1\sim 14$ letters. Three word examples are listed in Table (\ref{tab:ocrExamples}). The first two columns show two different handwritten patterns from different persons for the word listed in the third column. Apparently, each letter has different probability being transfered to other letters. As highlighted in Table (\ref{tab:ocrExamples}), letter `m' has high probability to be followed by `m', `a' or `b', while letter `n' will prefer to be transferred to `d', `g' or `i'. Intuitively, we suppose our proposed model will  take effect on this dataset and we verify it in the following.

\begin{table}
\caption{Examples of OCR dataset}
\begin{center}
\begin{tabular}{>{\centering\arraybackslash}m{0.97in}  >{\centering\arraybackslash}m{0.97in} >{\centering\arraybackslash}m{0.89in}}
\hline \\[-1.5ex]
\includegraphics[width=0.15\textwidth]{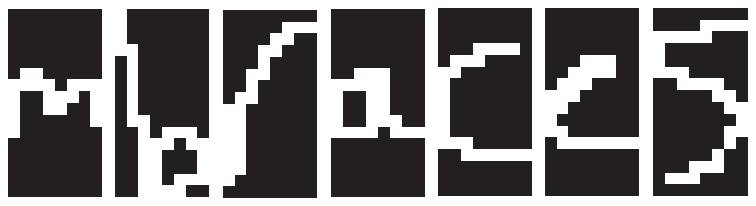} & \includegraphics[width=0.15\textwidth]{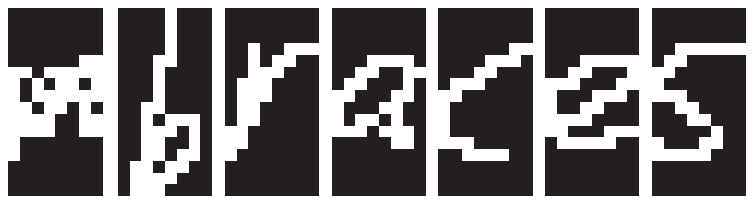} & \large{\textcolor{red}{m}braces} \\ \hline \\ [-1.5ex]
\includegraphics[width=0.15\textwidth]{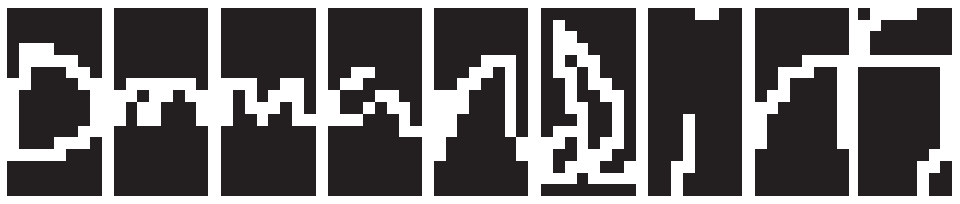} & \includegraphics[width=0.15\textwidth]{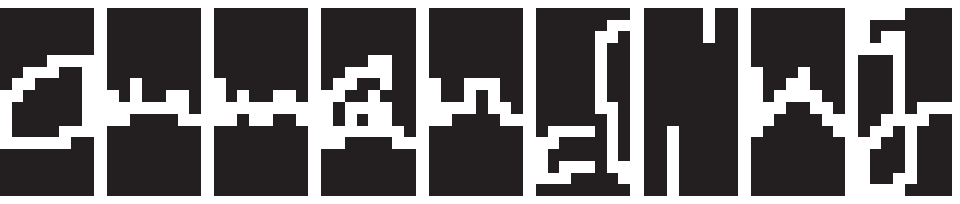} & \large{o\textcolor{red}{mm}a\textcolor{red}{n}di\textcolor{red}{n}g} \\ [1ex] \hline\\ [-1.5ex]
\includegraphics[width=0.15\textwidth]{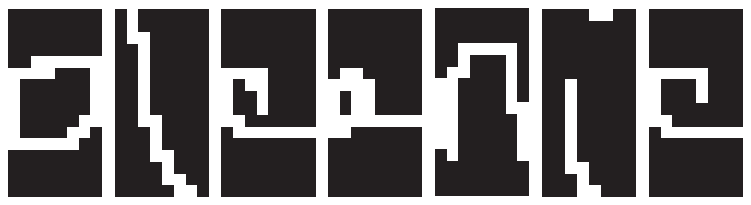} & \includegraphics[width=0.15\textwidth]{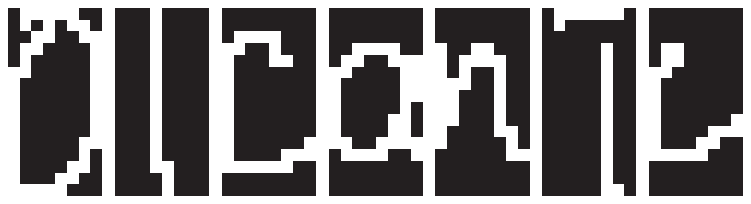} & \large{olca\textcolor{red}{n}ic}\\ \hline
\end{tabular}
\label{tab:ocrExamples}
\end{center}
\vspace{-2em}
\end{table}

We apply the true number of lowercase English letters as the size of the hidden state space, namely $k=26$. Accordingly, the initial state distribution $\pi$ is a $26$-d vector , and the transition matrix $A$ is a $26 \times 26$ matrix. Each observed letter image is reshaped into a binary $1\times 128$ vector. For the emission distributions, Naive Bayes assumption is applied and each dimension of binary vector is independently modeled by Bernoulli distribution, parameterized by $p_d, d\in {1,2,...,128}$, measuring the probability of that the current pixel value is equal to $1$. Finally, emission distributions $B$ is modeled by $26*128=3328$ parameters. In supervised setting, the parameters $\lambda = (\pi,A_0,B)$ are learned by MLE from the training set. All of our experiments are run with $10$-fold cross validation.

\begin{figure}
\center
\includegraphics[width=0.45\textwidth]{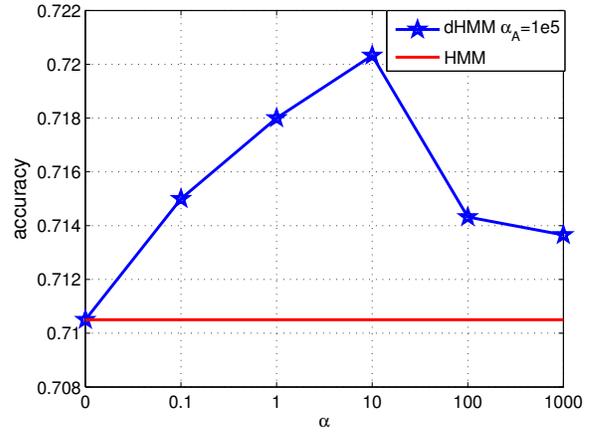}
\caption{Effectiveness of $\alpha$ for OCR}
\label{fig:OCR_alpha}
\end{figure}

Like PoS experiment, we first test the effectiveness of our proposed diversity-encouraging prior with a range of $\alpha$s. The test accuracies are shown in Fig. (\ref{fig:OCR_alpha}). The results are given by the averages across $10$ runs. Another parameter $\alpha_A$, which tries to drag $A$ to $A_0$, has been chosen through the 1-to-1 accuracy criterion and is fixed as $1e5$.  $\alpha=0$ corresponds to the traditional HMM and it gets an accuracy of $0.7102$ while our proposed dHMM obtains an accuracy of $0.7203$ with $\alpha=10$. That an increasing trend is gained demonstrates the effectiveness of dHMM though larger $\alpha$ will decrease the performance.

\begin{figure}
\center
\includegraphics[width=0.45\textwidth]{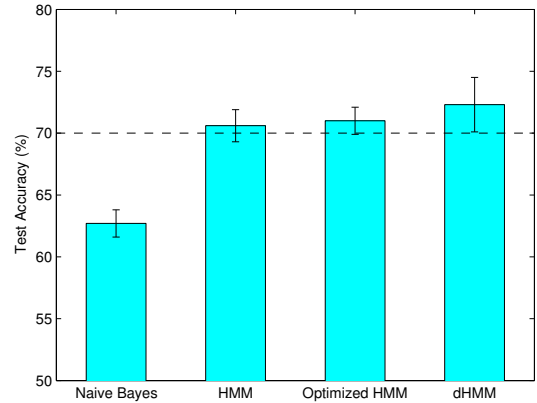}
\caption{Test accuracies of different classifiers}
\label{fig:ocr_accuracy}
\vspace{-2em}
\end{figure}

Our proposed model is compared to three baseline algorithms for supervised sequential labeling:  Naive Bayes, traditional HMM and Optimized HMM \cite{Elie2006}. The average accuracies with standard deviations are shown in Fig. (\ref{fig:ocr_accuracy}). Naive Bayes algorithm ignores the relationship between neighbor letters and achieves the lowest accuracy of $62.7\%$ with a standard deviation of $1.1\%$. Incorporating one-order chain structure of letters, HMM achieves a $70.6\%$ accuracy rate with a standard deviation of $1.3\%$. With other tricks, the optimized HMM obtains limited improvement. By contrast, by adding diversity-encouraging prior over the rows of transition matrix of traditional HMM, our proposed dHMM achieves an accuracy of $72.06$ with a standard deviation of $2.2\%$ which apparently gains a significant margin.

\begin{figure} 
\vspace{-1em}
\center
\subcaptionbox{{letter x}\label{fig:ocr_diversity_x} } 
 {\includegraphics[width=0.4\textwidth]{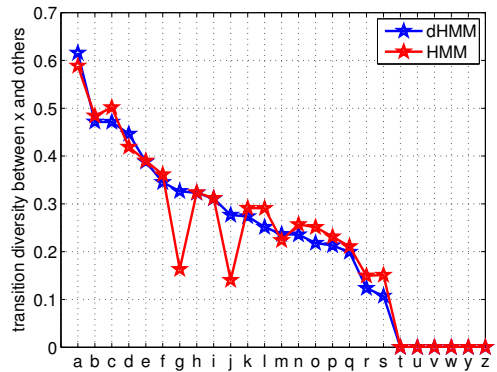}} \vspace{-1em}
\subcaptionbox{{letter y }\label{fig:ocr_diversity_y}}
{\includegraphics[width=0.4\textwidth]{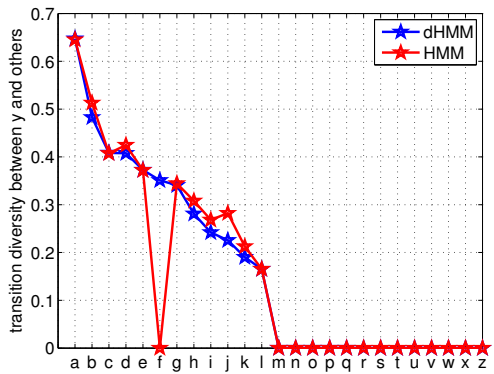}}
\caption{Transition diversity comparison between dHMM and HMM}
\label{fig:ocr_diversity}
\vspace{-1.5em}
\end{figure}

Finally, a qualitative demonstration of the diversity is shown in Fig. (\ref{fig:ocr_diversity}).
The transition matrix $A$ is trained from the setting of $\alpha = 10,~\alpha_A = 1e5$. Fig.(\ref{fig:ocr_diversity_x}) (Fig. (\ref{fig:ocr_diversity_y})) shows the diversity measurements (Bhattacharyya distance) between transition distribution of character `x' (`y') and transition distribution of the other $25$ letters. From the curves, the total trends almost are the same everywhere between traditional HMM and our proposed dHMM, except that dHMM heights the pairwise diversities between transition distributions of (`x',`g'), (`x',`j') and (`y',`f'), which we conclude contributes to the improvement of test accuracies in some extent .

\vspace{-1.2em}
\section{Conclusion}
\label{sec:conclusion}
Based on the methodology of traditional HMM, a diversified HMM (dHMM) for sequential labeling was proposed in this paper: Instead of explicitly constraining the parameters associated with the observations, we placed a diversity-encouraging prior over the parameters of transition distribution, modeled by Determinantal Point Processes (DPP), which is an essential part of traditional HMM. To facilitate this variation of HMM, a new Maximum A Posterior (MAP) scheme was proposed under both unsupervised setting and supervised setting. For unsupervised setting, maximum a posterior with marginal likelihood was solved based on the EM framework for the traditional HMM, but with a modified M-step. For supervised setting, maximum a posterior with joint likelihood was trained directly through the gradient descend method. We verified the effectiveness of our proposed dHMM through both the simulated and the real-world datasets (e.g., unsupervised PoS tagging and supervised OCR).

Our future work will involve with the development of a non-parametric extension to dHMM, which simultaneously learns the number of hidden states, as well as all HMM parameters. We will carry out a theoretical study into the effectiveness of the number of states as well as diversity-encouraging prior over rows of transition matrix under our setting with regard to labeling accuracy.
\vspace{-1em}

\ifCLASSOPTIONcompsoc
  \section*{Acknowledgments}
\else
  \section*{Acknowledgment}
\fi

The work was supported in part by the Australian Research Council Projects grants DP-120103730, DP-140102164, FT-130101457, and LP-140100569.

\vspace{-1.2em}
\ifCLASSOPTIONcaptionsoff
  \newpage
\fi
\bibliographystyle{abbrv}
\bibliography{bibtex}
\vspace{-3em}
\begin{IEEEbiography}[{\includegraphics[width=1in,height=1.25in,clip,keepaspectratio]{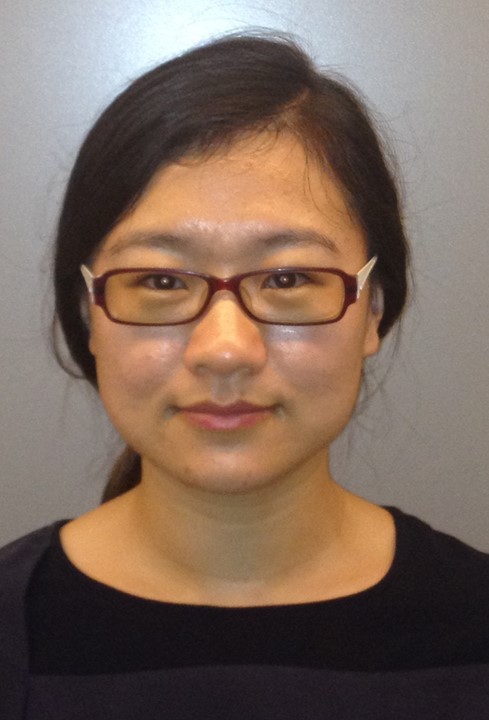}}]{Maoying Qiao}
received the B.Eng. degree in Information Science and Engineering from Central South University, Changsha, China, in 2009, and the M.Eng. degree in Computer Science from Shenzhen Institutes of Advanced Technology, Chinese Academy of Sciences, Shenzhen, China, in 2012. She is currently pursuing the Ph.D. degree with the University of  Technology at Sydney (UTS), Sydney, NSW, Australia.
Her current research topics are pattern recognition and probabilistic graphical modeling.
\end{IEEEbiography}
\vspace{-3em}
\begin{IEEEbiography}[{\includegraphics[width=1in,height=1.25in,clip,keepaspectratio]{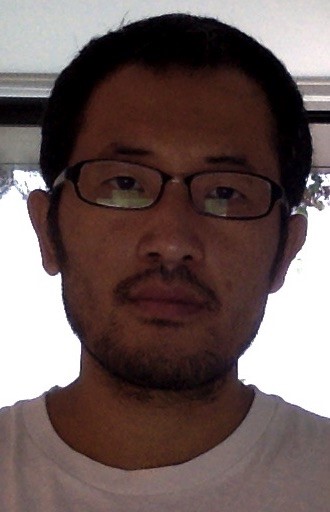}}]{Wei Bian}
(M'14) received the B.Eng. degree in electronic engineering and the B.Sc. degree in applied mathematics in 2005, the M.Eng. degree in electronic engineering in 2007, all from the Harbin institute of Technology, harbin, China, and the PhD degree in Computer Science in 2012 from the  University of Technology, Sydney, Australia.
His research interests are pattern recognition and machine learning.
\end{IEEEbiography}
\vspace{-3em}
\begin{IEEEbiography}[{\includegraphics[width=1in,height=1.25in,clip,keepaspectratio]{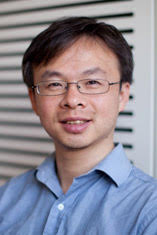}}]{Richard Yi Da Xu}
received the B.Eng. degree in computer engineering from the University of New South Wales, Sydney, NSW, Australia, in 2001, and the Ph.D. degree in computer sciences from the University of Technology at Sydney (UTS), Sydney, NSW, Australia, in 2006.
He is currently a Senior Lecturer with the School of Computing and Communications, UTS. His current research interests include machine learning, computer vision, and statistical data mining.
\end{IEEEbiography}
\vspace{-3em}
\begin{IEEEbiography}[{\includegraphics[width=1in,height=1.25in,clip,keepaspectratio]{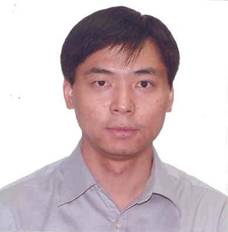}}]{Dacheng Tao}
 (F'15) is Professor of Computer Science with the Centre for Quantum Computation \& Intelligent Systems, and the Faculty of Engineering and Information Technology in the University of Technology, Sydney. He mainly applies statistics and mathematics to data analytics and his research interests spread across computer vision, data science, image processing, machine learning, neural networks and video surveillance. His research results have expounded in one monograph and 100+ publications at prestigious journals and prominent conferences, such as IEEE T-PAMI, T-NNLS, T-IP, JMLR, IJCV, NIPS, ICML, CVPR, ICCV, ECCV, AISTATS, ICDM; and ACM SIGKDD, with several best paper awards, such as the best theory/algorithm paper runner up award in IEEE ICDM'07, the best student paper award in IEEE ICDM'13, and the 2014 ICDM 10 Year Highest-Impact Paper Award.
\end{IEEEbiography}

\end{document}